\documentclass{article}
\usepackage{arxiv}
\usepackage[utf8]{inputenc} 
\usepackage[T1]{fontenc}    
\usepackage{hyperref}       
\usepackage{url}            
\usepackage{booktabs}       
\usepackage{amsfonts}       
\usepackage{nicefrac}       
\usepackage{microtype}      
\usepackage{lipsum}		
\usepackage{graphicx}
\usepackage{natbib}
\usepackage{doi}
\usepackage{tikz}
\usepackage{pgfplots}

\title{Replication of Multi-Agent Reinforcement Learning for the "Hide and Seek" Problem}
\date{ }	

\author{ \href{https://orcid.org/0000-0003-4205-2397}{\includegraphics[scale=0.06]{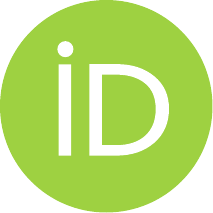}\hspace{1mm}Muhammad Haider Kamal}\\
	National University of Sciences and Technology\\
	Islamabad, Pakistan \\
	\texttt{haiderkamal@outlook.com} \\
	\And
	\href{https://orcid.org/0000-0002-2259-6824}{\includegraphics[scale=0.06]{Images/orcid.pdf}\hspace{1mm}Muaz Niazi}\thanks{Corresponding Author.
    \\ \emph{Email: drmak@bu.edu} (Muaz Niazi) } \\
	National University of Sciences and Technology\\
	Islamabad, Pakistan \\
	\texttt{drmak@bu.edu} \\
 \And
	\href{https://orcid.org/0000-0001-9583-5585}{\includegraphics[scale=0.06]{Images/orcid.pdf}\hspace{1mm}Hammad Afzal} \\
	National University of Sciences and Technology\\
	Islamabad, Pakistan \\
	\texttt{hammad.afzal@mcs.edu.pk} \\
}

\renewcommand{\headeright}{ }
\renewcommand{\undertitle}{ }
\hypersetup{
pdftitle={Replication of Multi-Agent Reinforcement Learning for the "Hide and Seek" Problem},
pdfsubject={q-bio.NC, q-bio.QM},
pdfauthor={Muhammad Haider Kamal},
pdfkeywords={Replicating Strategies · Multi-Agent Curriculum learning · Multi-Agent Competition · Drone competition},
}

\pgfplotsset{compat=1.18}
\begin{document}

\maketitle
\begin{abstract}
Reinforcement learning generates policies based on reward functions and hyperparameters. Slight changes in these can significantly affect results. The lack of documentation and reproducibility in Reinforcement learning research makes it difficult to replicate once-deduced strategies. While previous research has identified strategies using grounded maneuvers, there is limited work in more complex environments. The agents in this study are simulated similarly to Open Al’s hider and seek agents, in addition to a flying mechanism, enhancing their mobility, and expanding their range of possible actions and strategies. This added functionality improves the Hider agents to develop a chasing strategy from approximately 2 million steps to 1.6 million steps and hiders shelter strategy from approximately 25 million steps to 2.3 million steps while using a smaller batch size
of 3072 instead of 64000. We also discuss the importance of reward function design and deployment in a curriculum-based environment to encourage agents to learn basic skills along with the challenges in replicating these Reinforcement learning strategies. We demonstrated that the results of the reinforcement agent can be replicated in more complex environments and similar strategies are evolved including ”running and chasing” and ”fort building”
\end{abstract}


\section{Introduction}
Reinforcement learning generates policies based on designed reward functions and assigned hyper-parameters. Slight changes in these can yield differences in results. It can also be affected by experimental conditions, i.e., If the gravity of physics-based simulation or force applied to a working agent changes it can change the behavior pattern of the agent \cite{Baker2019}. \par
 
Replicating AI results and strategies is considered difficult in large part \cite{Ahadi2016}. Low replication rates suggest unreliable practices. Even detection of emergent behavior has historically been thought to be challenging \cite{Niazi}.  Thus, replication of once-deduced strategies is difficult and criticized for not being reproducible. Reinforcement learning research tends to be not documented well enough to reproduce the exact reported results as it mostly relies on continuous fine-tuning and updating of hyper-parameters, reward functions, environment variables, sensor types, etc \cite{Ahadi2016}.
Inference and result reproducibility should yield enough similar results that can benefit further research and can ease improvements rather than just focusing on reproducing it. But upon minimum info given regarding Agent’s Academy parameters (Environment Variables like gravity, collider conditions, delta time, etc.), Agent’s Behavior parameters (vector space, continuous/discrete actions inference device, collider type, speed, sensors attached, max steps allowed, etc.), calculating similar results is often considered hard to achieve\cite{Asplund2022}. 
\par
Strategies can either be generated by designing the reward function specified to goal-achieving parameters or waiting for the lucky shot action that the agent performs while moving randomly. The drawback of this approach is that the agent might never learn to reach the optimal solution if the experience that it is gathering is not fruitful \cite{Kelly2020}. For example, if the agent’s goal is vertically upward and behind a locked door, an agent might take forever to understand a pattern to first move toward a specific door, unlocking it, passing through it, and tagging the target goal.
 Thus, specially designed reward functions are deployed in a simplistic environment at the beginning to encourage the agent to learn a basic and novice version of the best state of action \cite{Price2020}.
 \par
 The paper Emergent Tool Use From Multi-Agent Autocurricula by \cite{Baker2019} presents a series of experiments in which agents are placed in a simulated environment and tasked with achieving a set of strategies. The agents are programmed to learn from their interactions with each other, leading to the development of strategies such as cooperation and tool use.
Our agents are simulated with the inclusion of a flying mechanism, enabling them to navigate through three-dimensional space. This feature shall enhance the agents' mobility and expand their range of possible actions, leading to more diverse and effective behavior. With the ability to fly, our agents can overcome obstacles and traverse complex environments more efficiently, ultimately improving their performance and increasing their chances of success. 
\subsection{Problem Statement}
\emph {“The Tool used from multi-Agent Auto curricula is currently trained to focus only Grounded Movements as there is no flying-like behavior (Drone) with no 3D observations along with curriculum learning.”}
\par
The current training approach for Tool use from multi-agent Auto curricula is limited to grounded movement due to the absence of a flying agent capable of 3D observations. This lack of aerial observation capabilities, coupled with the absence of curriculum learning, represents a significant challenge in replicating strategies and the development of more advanced and effective multi-agent strategies.
The research addresses the following research questions(RQ) in detail and proves.

\begin{itemize}
  \item  \textbf{RQ1} Does curriculum learning benefit hider agents for using tools?
  \item \textbf{RQ2} Does designing reward function specific to the goal accelerate the emergent strategy?
  \item \textbf{RQ3} Does constant negative reward damages Agent’s navigational phase?
  
\end{itemize}

\section{Method}
\label{Method}
The research was designed keeping in mind the Open AI’s Hide \& Seek experiment with the addition of flying-like drone mechanics to enable pure 3D maneuver in a more complex environment. Our agents are trained using reinforcement algorithms. PPO for Hiders and MA-POCA for seekers. We also used curriculum learning and self-play. We compared the environmental steps and episode length of resulting strategies when they emerged. The training was divided into two sections. prep phase and test phase.
\subsection{Multi-Agent Prep-Phase}
Agents are given a total of 3072 steps in which they can make decisions every third step and update their actions accordingly. Hiders are given a 40\% preparation time before the seekers are allowed to move. In this “prep phase” hiders are allowed to move freely while avoiding contact with seekers. Hiders can learn to identify, drag, and lock movable props (i.e., boxes) to use them to their advantage. During this period, seekers are observing the hider’s location and shapes (if hiders accidentally stay in front of the seeker’s field of vision) but they are restricted with their movement.
\begin{equation}
S_p= S_m*(40/100)
\end{equation}
Where $S_p$ are the Steps for the prep phase and $S_m$ are the Max Steps.

 \subsection{Multi-Agent Test-Phase}
The test phase refers to that portion of training in which the seekers are let loose and given control over themselves. That means seekers can now also move along with hider agents and try to hunt them and tag them. After 40\% of steps are completed out of max environment steps, for the rest 60\% of the steps, seekers are allowed to move and use their learned policy to focus on targets and reach them.
\begin{equation}
S_t= S_m*(60/100)
\end{equation}
Where $S_t$ are the Steps for the test phase and $S_m$ are the Max Steps.

\subsection{Observing Player Progression via Curriculum learning}
Our Hider agents are trained via a form of machine learning method known as curriculum learning which includes progressively raising the level of complexity of the training examples provided to an AI agent. Task Sequencing: The first step in curriculum learning is to define a sequence of tasks with increasing difficulty. This sequence can be represented as a function that maps the current training iteration to the corresponding task. For example, the task at iteration t could be represented as:
\begin{equation}
f(t)=Task_t
\end{equation}
 The plan is to start with simple instances and progressively get more complicated as the agent gets better at the task.
\begin{equation}
C_r> T_r \to D +=1
\end{equation}
Where $C_r$ is the current reward, $C_r$ is the Cumulative threshold reward. $D$ is the difficulty as in levels. That is the current reward exceeds a given number of episodes, the environment will evolve to complex itself to make it harder for an agent to learn gradually. Agents are introduced to four different levels of environmental setup. Levels are made difficult to complete progressively. For example, level one has the least difficulty while level four has the maximum.  Curriculum learning is applied to educate the agents on how to play the game of hide and seek more effectively. The hiders and the seekers are two different teams of agents in this game. The hiders must conceal themselves in their surroundings, while the seekers must track them down and tag them. By gradually escalating the level of difficulty, curriculum learning is integrated into the game of hide and seek with the students. The hiding places could be harder to locate as the hiders get more adept at doing so. Similar to this, as the game goes the curriculum learning in ml-agents hide and seek is intended to assist the agents in learning more quickly and effectively. The agents can learn the game in a more organized and effective manner by starting with simple instances and progressively adding complexity.

\subsection{Seeker's Assault Horizon}
The idea behind this is inspired by the military term “Dog Fight”. A dogfight is a kind of aerial conflict in which two or more aircraft engage in a close-quarters battle. High-speed maneuvers and intricate strategies are frequently used, with each pilot attempting to outwit their rival. Dogfights may be quite dangerous since they frequently take place at great heights and make use of cutting-edge equipment. Our agent’s navigational capabilities are designed by keeping the “running and chasing” ability in mind. Seekers are to keep hiders in their assault horizon to gain rewards and even tag them to get an additional reward. 
With the help of a specially designed target field of vision “reward strength” signal, seekers can identify what angle suits best to adjust its rotation and position accordingly, so it is facing and is right in front of its target.
Let $T$ = forward vector “z-axis” i.e. (0,0,1) and let a = difference between this agent’s position  $P_1$ and other agent’s position $P_2$.
\begin{equation}
a= P_2-P_1
\end{equation}
\begin{equation}
R_d= |a|
\end{equation}
\begin{equation}
d=T.(a/R_d )
\end{equation}
\begin{equation}
S_s=(d/R_d )
\end{equation}
Where $S_s$ is the Signal Strength, $R_d$is the magnitude of $a$, $d$ represents the dot product between the forward vector and $(a/R_d )$

\subsection{Prop increases via curriculum}
As our agents have flying-like capabilities, introducing a ramp was of no use as agents might learn to just fly over the wall. So, we introduced windows and doorways which when blocked with props, make a safe house for the hiders. There is a single prop assigned for each window or door. 
Let $P_n$ represent the number of props, $L_n$ represent the level number, and R represent the reward number. Then, we can represent the given statement in mathematical notation as:
\begin{equation}
P_n= f(L_n,R)
\end{equation}

This equation states that the number of props $P_n$ is a function f of the level number $L_n$ and the reward number $R$. 
Agent must learn to efficiently handle the props to minimize the time required to close that door and learn which prop is best for quickly closing a specific location door. Agent shall be given a reward for closing one door/ window successfully. If the agent is using the farthest prop to close a door It will damage the efficiency as it needs to close the rest of the doors to clear obstacles

\section{Results and Analysis}
In the previous section, the methodology of our two agents including the algorithms, observations, and their reward distribution has been described. The following section includes the results and analysis of Hiders and Seekers agents. This section will be divided into two major portions each highlighting the resulting behavior and analytical outcome of Seekers and Hiders.
By using the methodology explained in the above section it’s been deduced that training agents with a lesser negative reward or lesser penalties yield a quicker learning behavior and can result in faster convergence to the optimal policy. This answers our \textbf{RQ3}. This occurs especially in navigation-related tasks where an agent might be in the right direction to target but is getting a penalty for not looking at it. PPO and MA-Poca tend to increase their rewards based on observations, so it is beneficial to leave the negative rewards at a minimum. Hiders are given a 0.001 reward for every frame they are hidden while seekers are given a 0.001 reward if hiders are in the field of vision and +1 if they crash hiders.
\newline
Seekers are trained in a group of 2-4 multi-agents and assigned cumulative rewards i.e., each will get the reward for good action, and each will get a penalty if one does a bad action. It resulted in the occurrence of cooperation among them. Seeker’s behavior shows they tend to, with time, explore random paths less and use the target’s (Hider) location observation to calculate the shortest path and shortest facing direction. This behavior helps the seekers to put hiders easily in their FOV (field of vision). Seekers were successful in emerging “Running and chasing” which includes sub-strategies like locating, navigating, finding Hiders, navigating around walls, and door identification. As Seekers are already flying (drones) they learned to ramp themselves up with an upward thrust through the windows. As windows do not allow the drone to enter without moving vertically up or down (‘down’, in our case as the agent can be near the roof) the “ramp use” strategy is also satisfied as they are successful in entering the hider’s rooms via learning vertical movement and clearing obstacles Infront of windows.
\newline
Hiders are trained solo and as a group of 2 multi-agents. As their main reward is based on isolating themselves in a safe location their rewards are quite far in the future i.e. agents need to perform more random actions to identify the best state-action situation to push in the buffer to generate an optimal policy. To overcome this barrier, we used curriculum learning which by increasing the complexity of the environment gradually yields an agent understanding of the basic concept of what actions are required to get a maximum reward in an episode. 
\newline
The environment has six key components. Walls, Props, Hiders, Seekers, Boundaries, and obstacles. Walls are static objects designed specifically to gradually increase the difficulty by enabling or disabling a chunk of it as shown in figures 1-4 above. Props are used as a door or a shield that can be dragged into the vacant door/window spaces and locked there so seekers are unable to get inside the hider’s safe hideout. While obstacles are pure physical rigid bodies that can be pushed. They have gravity involved with much less weight which ends in gliding behavior that continues to stay in the path of agents even after being pushed once. By inducing such curriculum learning behavior, hider agents tend to grab the concept of safely dragging the prop to the opening and closing it more quickly and efficiently than directly starting the training at level four. One can reduce or increase the levels just to get optimal results from the agents. This subtle increase in the complexity of the environment rather than forcing the agent to adapt to the more complex one yields beneficial results. We did two experiments by training the hider agents until 3.3M steps in both traditional and curriculum learning. In comparison to traditional non-curriculum learning experiments, there was no increase in reward cumulative rewards as shown in the graph Figure 1.
\subsection{Curriculum Learning vs Traditional Learning}
\begin{figure}[htb]
 \begin{center}
    \includegraphics[width=0.9\columnwidth]{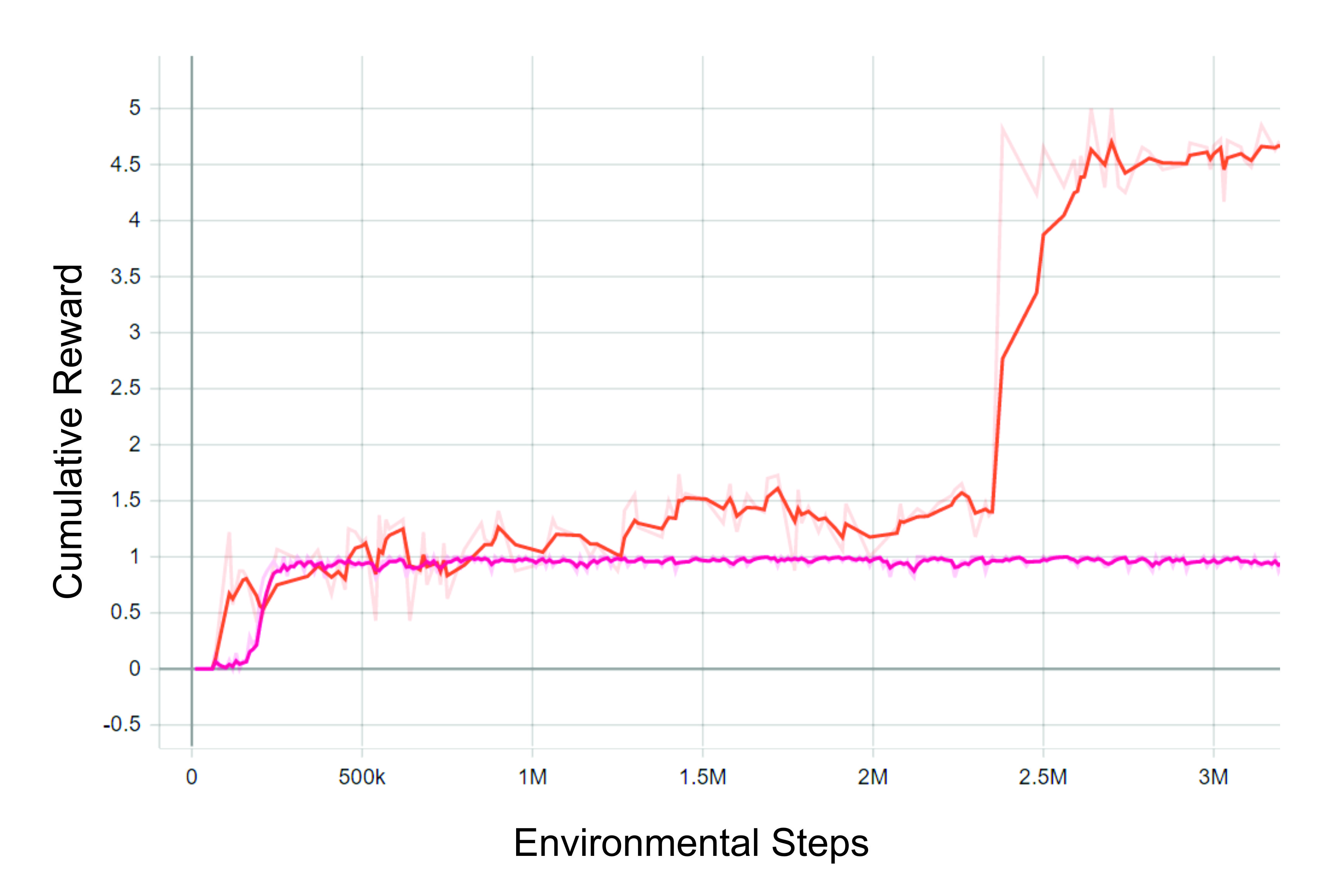}
\end{center} 

\caption[Curriculum vs Traditional]{Reward Distribution for Hider Agents comparing curriculum vs traditional learning setup.}
\end{figure}
 
 {\par \small
Our experiment also represents cumulative reward comparison over 3.3M environmental steps between curriculum vs traditional learning environmental setup. This answers the \textbf{RQ1}. The solid magenta-colored line represents the traditional approach where Hider agents were able to just run away from seekers satisfying the first strategy and not yielding to the second. The solid orange line represents curriculum learning where training was divided in the forms of levels which yields better continuous reward per frame as hiders were able to hide for more time steps and get more reward every frame.}
\newline
Reward, in the case of hiders, is given on the basis of how many frames the hiders can hide from the field of vision sensor of seeker agents. More the frames they are hidden the more they can get rewarded. With the curriculum, learning involved hiders were able to learn a basic behavior and adopt a converging policy that resulted in a better hiding behavior. Therefore, the curriculum learning approach was adopted for further experimentation.
 \newpage
\subsection{Hiders}
The following section discusses and analyses the experiments and their results according to hider agents.
\subsubsection{Cumulative Reward}
This section covers Hider's group cumulative reward.
 \begin{figure}[htb]
  \begin{center}
\includegraphics[width=1\columnwidth]{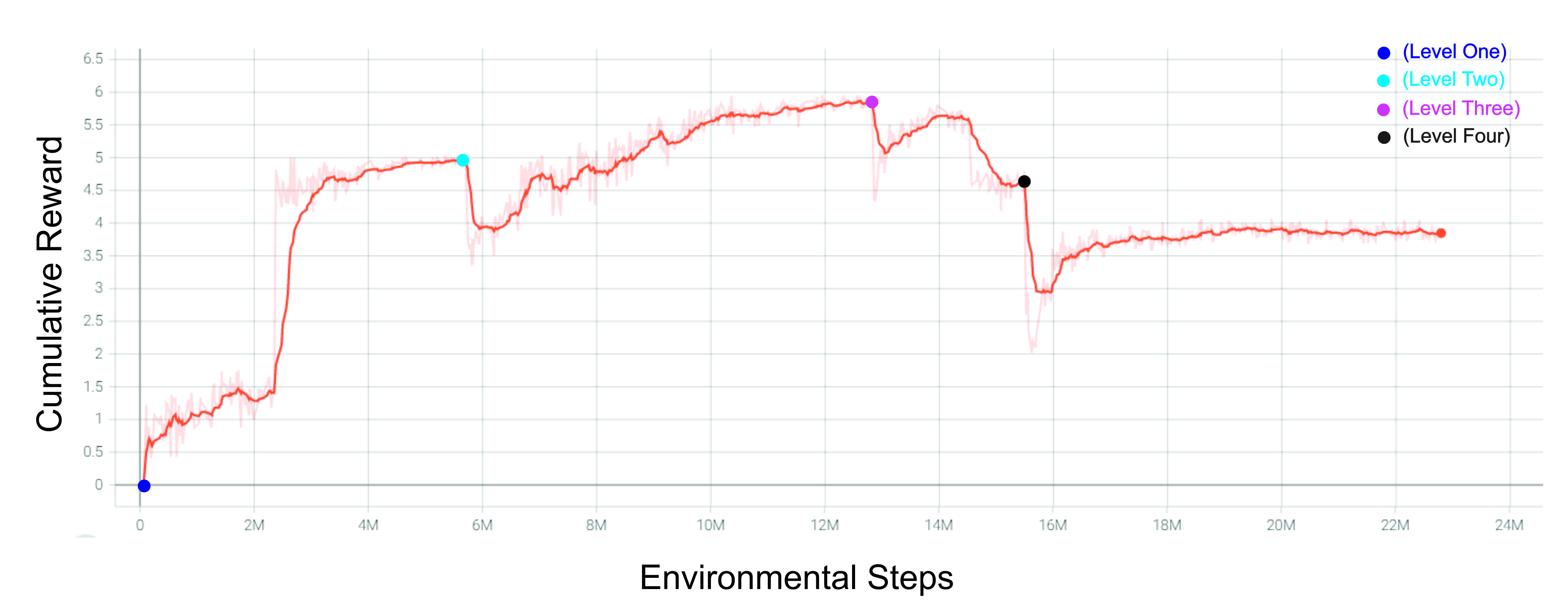}  
 \end{center}
\caption[Hiders Cumulative Reward]{Hiders Cumulative Reward Smoothed 0.99.}
\end{figure}
 
  {\par \small The solid orange line represents Hider’s cumulative reward over the environmental steps. Training is initialized using Level One. Level Two is introduced at 5.7M environmental steps. Level Three at 12.8 steps while Level Four with decreased reward per frame at 15.7M steps}
 \newline
A spike at 160k environmental steps axis represents Hider Agents were able to learn to get some reward by crudely running and escaping from the seekers. Until 2.3M steps, Hiders were running and escaping from the seekers more efficiently and thus were getting more rewards per episode. From 2.3M to 5.7M a sudden increase in reward was observed as level one (Figure 19) of curriculum learning was finally working and yielding maximum reward. At the 5.7M step, level two was introduced and agents experienced a sharp decrease in rewards as the difficulty was increased. But as they have already a semi-trained brain, they continued to update the policy and adopt it. At the 12.8M step, the third level was introduced which shows a steep downfall indicating complexity for the agent to learn. Finally, at the 15.2M step level, four was introduced and the reward was also reduced for an agent to learn from the very complex condition, but it still managed to maintain a solid positive reward after 16.5M steps.
 
\subsubsection{Episode Length}

 \begin{figure}[htb]
 \includegraphics[width=1\columnwidth]{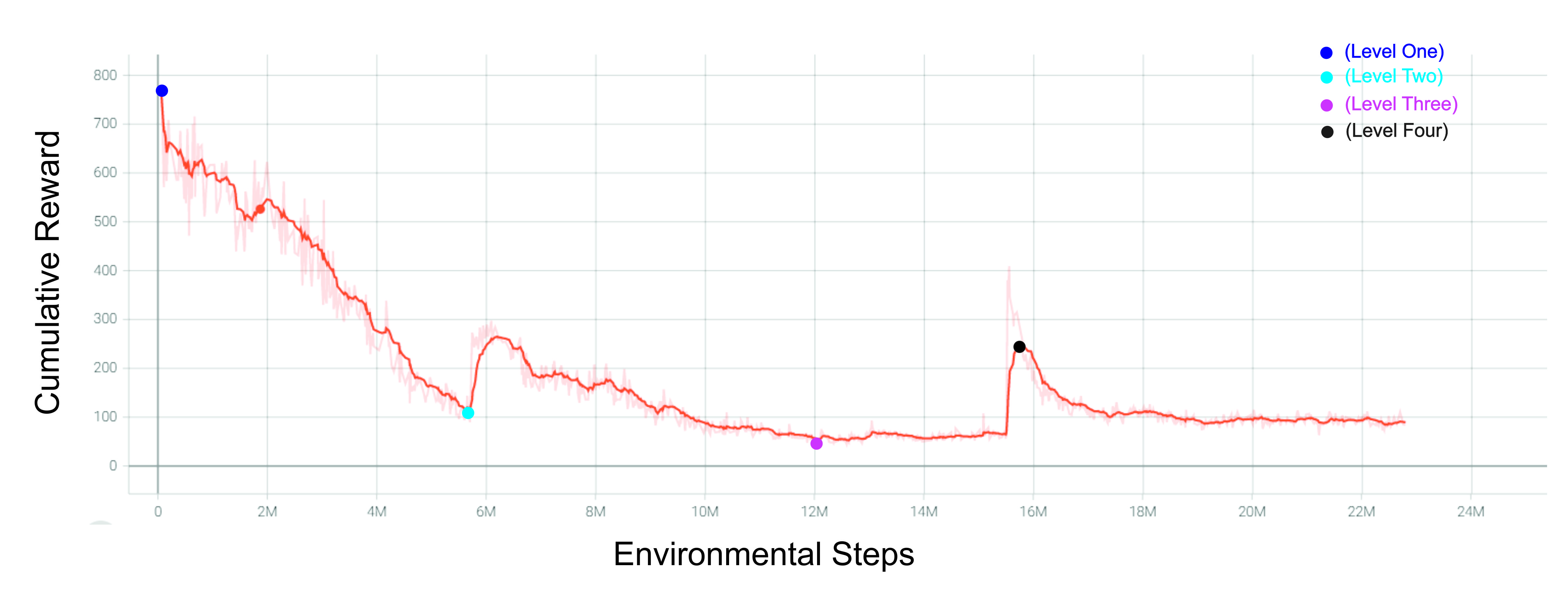} 
\caption[Episode Length - Hiders]{Episode Length of Hiders Corresponds to decisions made per episode.}
\end{figure}
 
  {\par \small The graph represents the gradual decrease in episode length as few decisions are required to complete an episode. The solid orange line represents the episode length over the environmental steps.
Training starts with 768 decisions made per episode
}
\newline
The training environment is designed in such a way that it will reset the episode as soon as one of the agents is successful in making the next strategy. i.e., if the “running/chasing” hider performs “fort building”., the environment will reset to promote the use of lesser episode length. This behavior encourages the agents to find the optimal next policy and strategy faster. Observing the graph Figure 3, we can see a slight increase in episode length at 5.7M steps where level two of the environment is introduced. This means the agent requires more steps to find a new policy to get more rewards. At the 15.8M step, another bump is observed showing level three is deployed and it affected the current behavior such that it needs more time to converge. After 18M steps, there seems a steady graph representing a steady and constant episode length. This means agents need at least 21 steps to converge to the optimal policy. 
\newpage
\subsection{Seekers}
The following section discusses and analyzes the experiments and their results according to hider agents.
\subsubsection{Group Cumulative Reward}
 \begin{figure}[htb]
 \includegraphics[width=1\columnwidth]{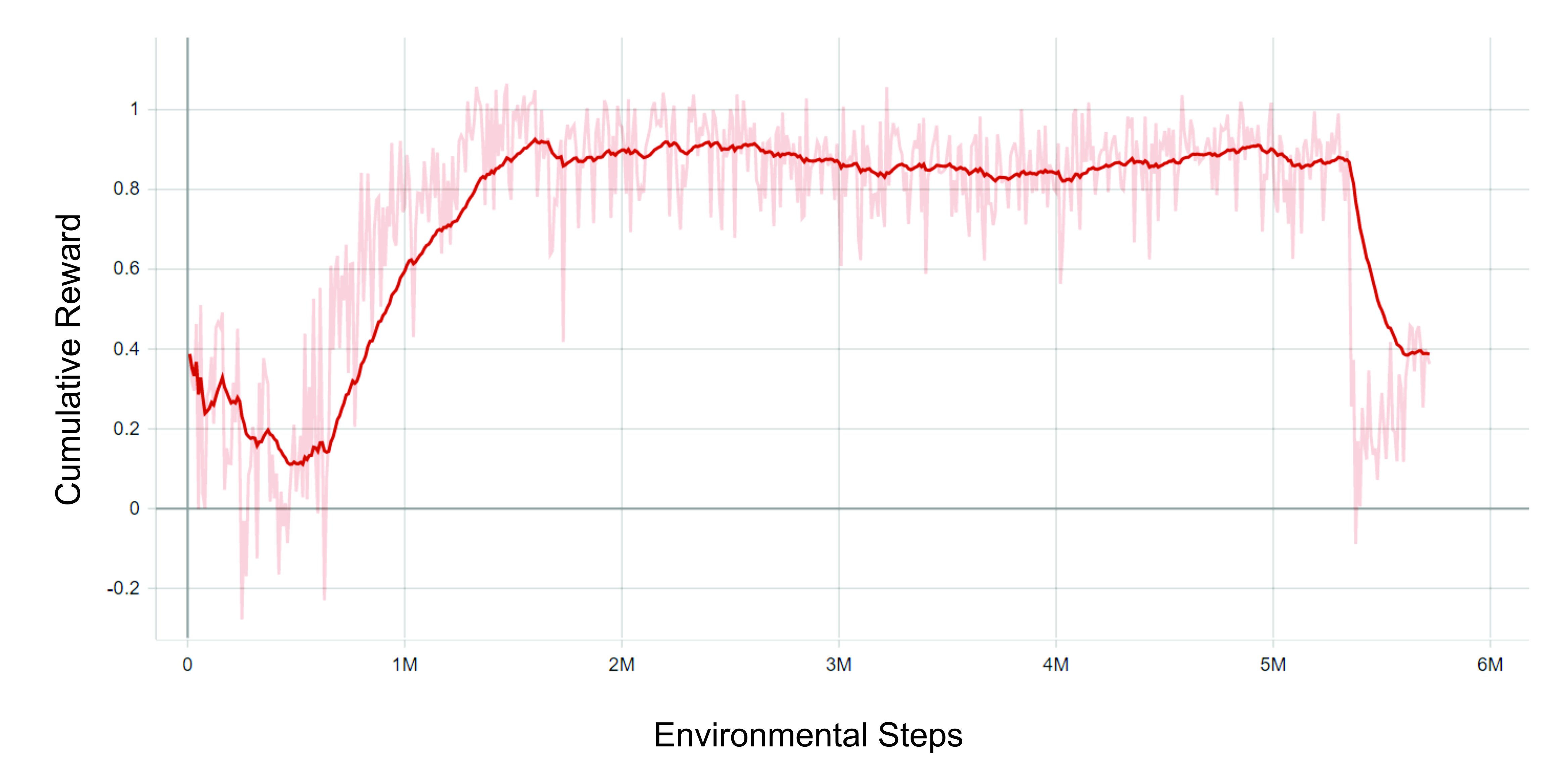} 
\caption[Seeker's Group Cumulative Reward]{Seeker's Group Cumulative Reward Smoothed 0.95}
\end{figure}
 
  {\par \small Training is initialized using Level Four. At 630k steps, seekers learned to rotate toward the hiders and keep them in the field of vision.  At 1.6M step seekers were successful in traversing obstacles, navigating through walls, and moving toward hiders. At 5.3M step seekers were trained again just to check how they behave when hiders are also trained for level four environmental setup, and it shows a steep descent in reward as they were not unable to see hiders.  The solid red line represents the group cumulative reward given over environmental steps.
}
\newline
Seekers are trained in two phases, first one is from 0-5.3M steps. This phase includes the “running \& chasing” hiders as a competition so they will learn this strategy as well. Phase two starts at 5.3M steps which includes training with “fort building” hiders that yield a sharp descent in the graph. This indicates the complexity and difficulty the seekers were facing in locating the hiders once they emerged to the second strategy. Seekers are given a +0.001 reward for every frame they see hiders in their field of vision. And no reward when they do not have them in sight. This is done because if we give a negative reward for such a constant frame rate just because the seeker’s field of vision does not contain hiders, we are damaging the path navigation behavior. Such that an agent might be on the right path but is getting penalized just because it is not there yet. Seekers are tested in a group of 2-4 multi-agent setup and solo as well. While training the 2-seeker agents with the same brains are being trained. Dual input is used but the rewards are given on a team basis. If one agent is successful both agents will get rewarded and will share the updated best policy while if one is getting penalized both will face this.
\subsubsection{Episode Length}
\begin{figure}[htb]
  \includegraphics[width=1\columnwidth]{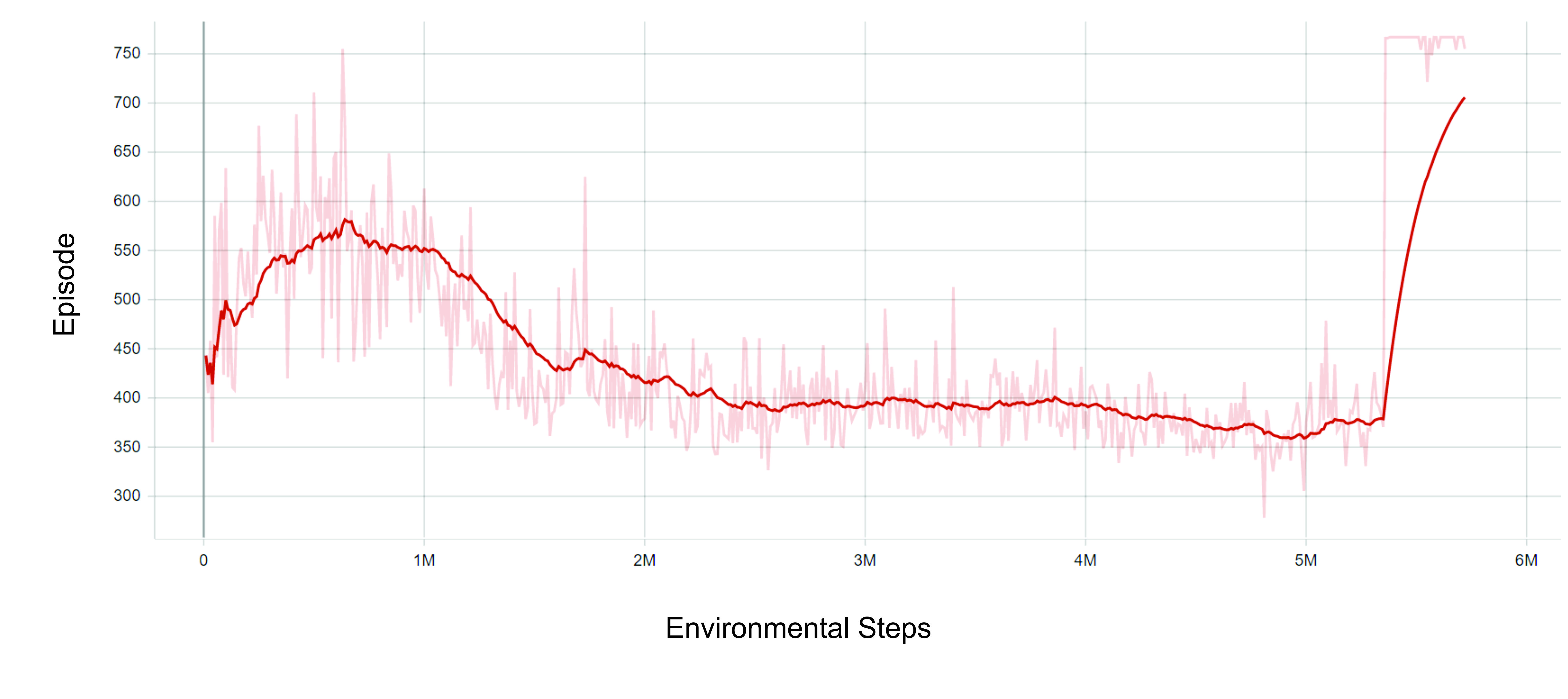} 

\caption[Episode Length - Seekers]{Episode Length for Seeker agents Corresponds to decisions made per episode}
\end{figure}
 
{\par \small The graph represents the gradual decrease in episode length as few decisions are required to complete an episode and a sudden increase demonstrating agents require more decisions when the opponent is far more difficult. The solid red line represents the episode length over the environmental steps.
}Seeker agents are programmed in such a way that if they have hider agents in sight, they get 0.001 rewards per frame but if they collide with them the episode ends. Thus, the more the seekers can collide quickly the lesser the episode length gets. This is done to encourage quick navigation and faster resetting of the training environment because if the seeker can collide with the hider, the hiders are not protected well, and there is less benefit in training again in an already won episode. 

\newpage
\subsection{Key Difference }
As Larger batch sizes require more GPU memory and processing power \cite{Rotenberg}, with our approach and methodology, we proved that agents emerged strategies much faster and efficiently with much less processing power. While batch size is equal to number of steps in an episode, buffer size was selected using the following formulae:

\begin{equation}
    B=b* E_n* P_n
\end{equation}
Where:
$B$ = Buffer Size
$b$ = Batch Size
$E_n$ = Environment Count
$P_n$ = Instances Count

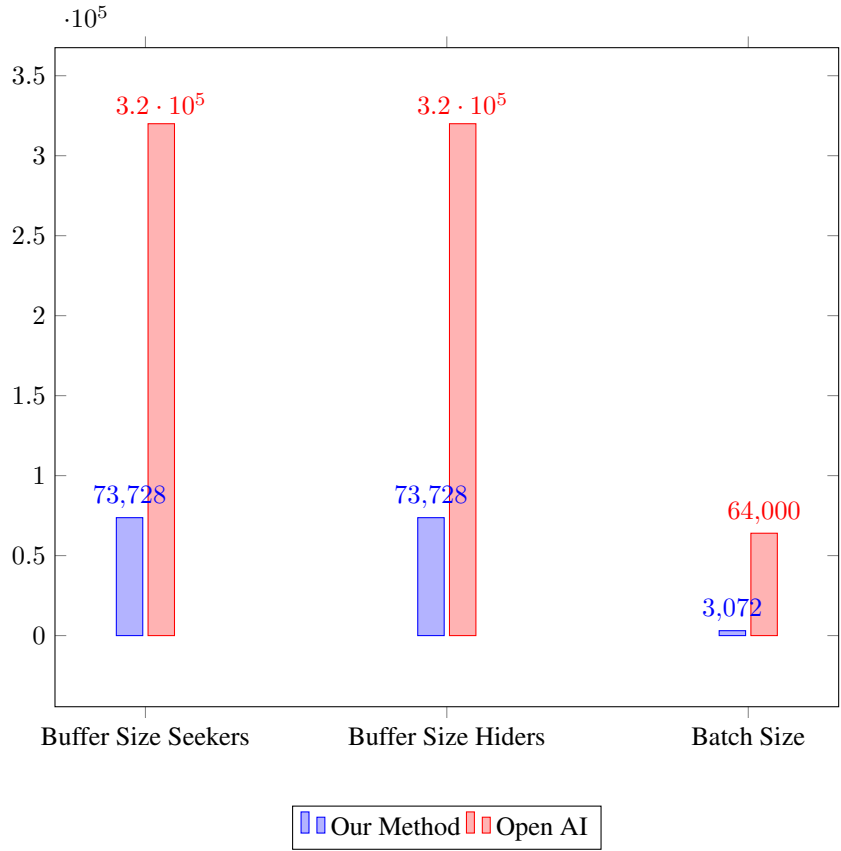
\begin{figure} 
\begin{tikzpicture}  
\pgfplotsset{width=12cm,compat=1.18}  
\begin{axis}  
[  
    ybar,enlargelimits=0.15,legend style={at={(0.5,-0.15)}, anchor=north,legend columns=-1},
    ylabel={},
    symbolic x coords={Buffer Size Seekers, Buffer Size Hiders, Batch Size},  
    xtick=data,  
    nodes near coords,  
    nodes near coords align={vertical},  
    ]  
\addplot coordinates {(Buffer Size Seekers, 73728) (Buffer Size Hiders, 73728)(Batch Size, 3072)}; 
\addplot coordinates {(Buffer Size Seekers, 320000) (Buffer Size Hiders, 320000)(Batch Size, 64000)};

\legend{Our Method, Open AI}  
  
\end{axis}  
\end{tikzpicture}     

\caption{Compression of our approach and Open AI's approach }
\end{figure}
 
Our method utilizes significantly fewer system resources compared to Open AI's experiment for deriving strategies as the buffer size required by our approach is approximately one-fourth of what was needed in the Open AI experiment. Similarly, the batch size required by our model is approximately one-twentieth of what was required by Open AI. This significant reduction in resource requirements demonstrates the efficiency and effectiveness of our approach.
\begin{center}
    
\begin{figure} 
\begin{tikzpicture}  
\pgfplotsset{width=10cm,compat=1.18}  
\begin{axis}  
[  
    ybar,enlargelimits=0.15,legend style={at={(0.5,-0.15)}, anchor=north,legend columns=-1},
    ylabel={},
    symbolic x coords={Seeker’s chasing strategy, Hider’s shelter strategy},  
    xtick=data, 
    nodes near coords,  
    nodes near coords align={vertical},  
    ]  
\addplot coordinates {(Seeker’s chasing strategy, 1600000
) (Hider’s shelter strategy, 2300000
) }; 
\addplot coordinates {(Seeker’s chasing strategy, 2000000
) (Hider’s shelter strategy, 25000000
) };

\legend{Our Method, Open AI}  
  
\end{axis}  
\end{tikzpicture} 
\caption{Compression of Emerged Strategies }
\end{figure}
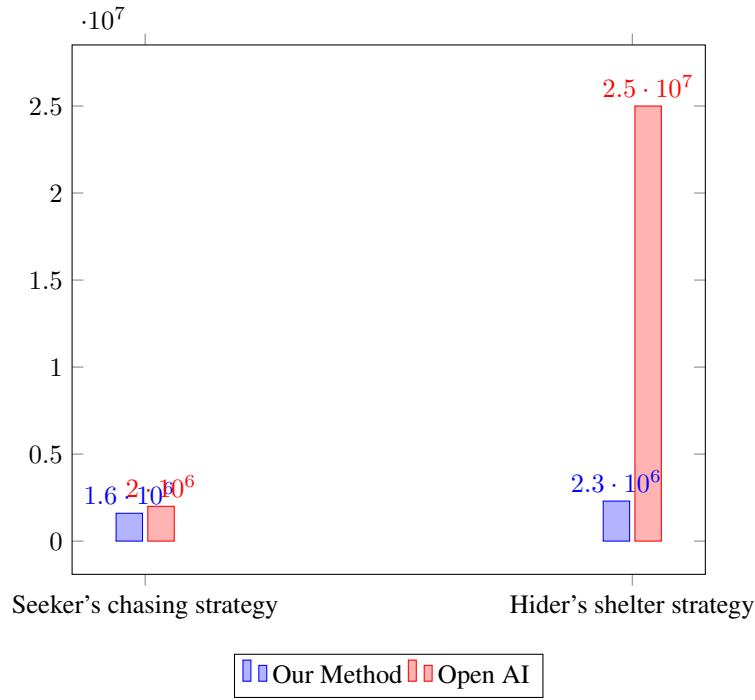
\end{center}

In addition to requiring fewer resources, our approach also deduces strategies in a much earlier environmental step count. For instance, the Running and Chasing strategy was identified approximately 0.4 million steps earlier in our approach than in the Open AI experiment. Similarly, our model was able to discover the strategy for creating shelter 22.7 million steps earlier than in the Open AI experiment. This early identification of strategies could prove to be a significant advantage in certain contexts. This answers our \textbf{RQ2}.
Overall, our approach offers a more resource-efficient and time-efficient method for deducing strategies than Open AI's experiment. By requiring significantly fewer resources and identifying strategies earlier in the environmental step count, our model could prove to be an effective tool in various domains. The results of our experiment demonstrate the potential of our approach to advance the field of strategy identification and inform decision-making in a range of practical applications.
\section{Conclusions and Future work}

In conclusion, our research has shown that replication of strategies is possible by utilizing a combination of simple game rules, fostering multi-agent competition, and implementing standard reinforcement learning algorithms at scale, and agents can acquire complex strategies and skills. The introduction of curriculum learning has further accelerated the learning process, allowing agents to utilize their environments and props more efficiently. This approach has the potential to significantly reduce costs and improve efficiency in real-world environments, particularly in the field of autonomous systems.
Furthermore, the development of autonomous systems has been a crucial area of research in recent years. The ability to deploy autonomous systems in a range of industries has the potential to improve safety, reduce costs, and increase efficiency. Our research has contributed to this field by providing a framework for the development of intelligent agents capable of learning complex strategies and skills.
Overall, the potential applications of this research are vast, particularly in industries such as transportation, logistics, and agriculture. By utilizing intelligent agents capable of learning complex behaviors, these industries could reduce costs, improve efficiency, and enhance safety. Therefore, our research has the potential to make a significant impact on the development of autonomous systems and has promising implications for a range of industries.
The training of agents can be extended to encompass more complex environments, where additional agent types can be introduced to heighten the level of training difficulty. The incorporation of weather parameters can also enable agents to adapt to challenging weather conditions, thereby enhancing their overall performance. However, it has been observed that agents may exploit the physics of their environment to manifest glitched, cheat-like behaviors. For instance, seekers have been observed to glitch through walls when hit with high velocity, while hiders tend to immobilize themselves in corners of the boundary to avoid incurring a significant collision penalty from seekers. The elimination of such glitched behaviors through the design of an environment that precludes their manifestation can dramatically reduce the time and resources expended in the agent training phase. As such, the creation of an environment that discourages such undesirable behaviors is a crucial aspect of effective agent training.

\nocite{*}
\bibliographystyle{unsrtnat}
\bibliography{manuscript}   

\section{Appendix}
\subsection{Agent Structure}
 \includegraphics*[width=5in, height=4in]{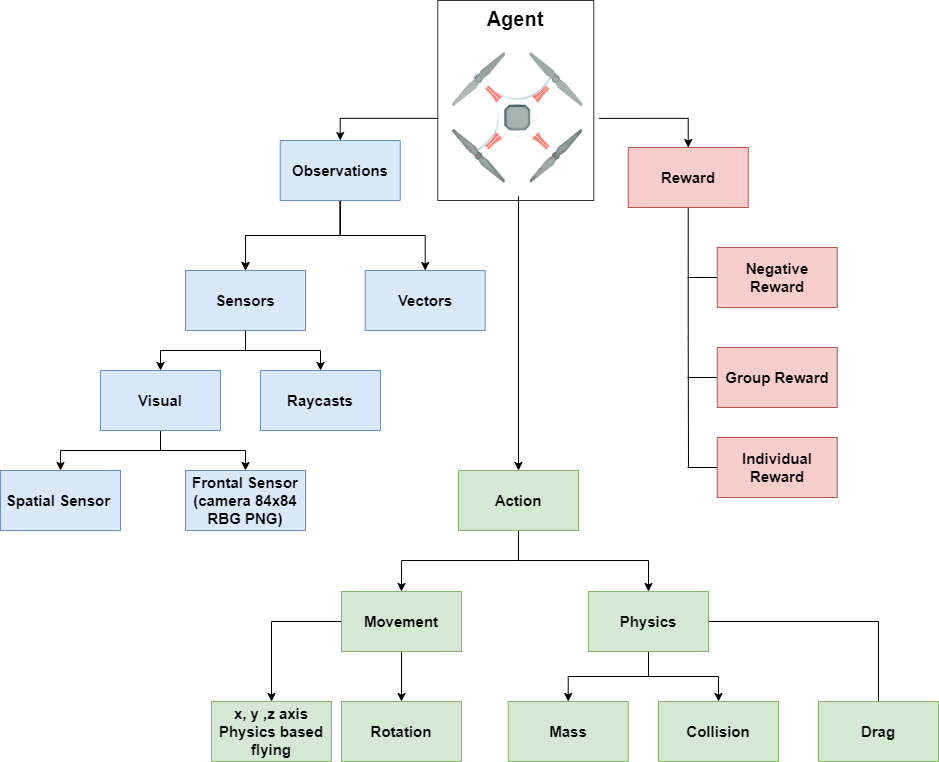}
\begin{center}
  {Figure 8 High-level architectural design of Agent}\textbf{ {}}
\end{center}

Figure 8\textbf{ }represents the architectural structure of an agent. The types of observations it receives, the reward types it gets, and the pilot parameters that include movement and physics-related types.

To explain further, the upcoming section of this chapter includes descriptions regarding the following:
\subsubsection{Agent:}
 
An agent is characteristically represented as a neural network that takes in Observational inputs from its environment and outputs actions. The agent learns to perform these actions in response to its environment by training on a dataset of observations and rewards. The following are the three observation types our agents are receiving which are thoroughly explained further in this chapter:

\begin{center}
\begin{tabular}{|p{1.3in}|p{1.3in}|p{1.3in}|} \hline 
\textbf{Spatial Sensor} & \textbf{Frontal Sensor} & \textbf{Raycasts} \\ \hline 
\includegraphics*[width=1.37in, height=1.40in]{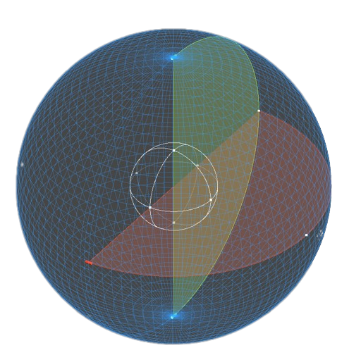}\newline  {Figure 9 ``360'' in 3d space} & \includegraphics*[width=1.37in, height=1.41in]{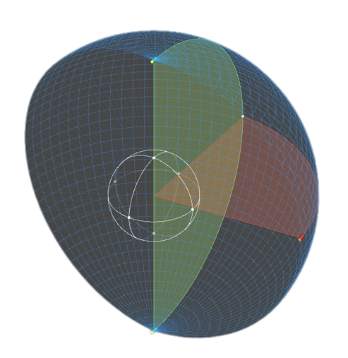}\newline  {Figure 10 ``135'' in 3d space\newline } & \includegraphics*[width=1.2in, height=1.in]{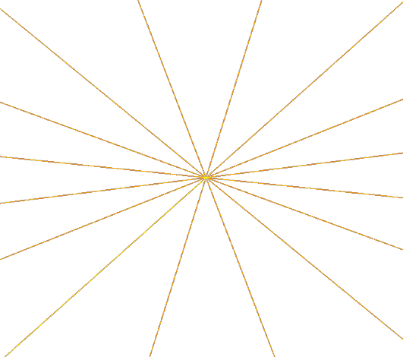}\newline  {Figure 11 "360" in 2d space\newline } \\ \hline 
\end{tabular}
\end{center}
 
\subsubsection{Hider Agents:}
One of our agents, whose objective is to flee from the seeker agents and build a shelter out of the environment's objects. These agents are rewarded when they are successful in hiding from seekers either by staying out of their sight or by building a fort for more security.

\subsubsection{Seeker Agents:}
The other type of our agents whose objective is to find and tag hider agents. They must learn to navigate toward the hiders and collide with them. Seeker agents get rewards is hider agents are in their sight or they tag any hider. These seekers are responsible for learning movement in flying like drone maneuvers, identifying the target (Hiders), locating that target, adjusting its facing direction to the target, and moving toward it avoiding obstacles and static walls.
 
\subsubsection{Neural Network structure:}
Agents are assigned 256 hidden units along with 2 hidden layers. 

\subsubsection{State Abstraction}
Spatial Grid Sensors are designed for capturing agents in 3D space. It contains a set of sensors that capture the state of the environment by dividing the space into a 3D grid of cells and recording the occupancy or other features of each cell.

\subsection{Grid Construction: }
To construct the grid, we divide the 3D space into a set of uniformly sized cells. Each cell is defined by its center point, which can be calculated as:
\[C_i=\ B_i+\left(\frac{1}{2}\right)*S_i\] 

Where:

 $C_i$ is the center point of the i-th cell.
 $B_i$ is the bottom-left corner of the i-th cell.
 $S_i$ is the size of the i-th cell.

\subsubsection{Occupancy Calculation: }
The occupancy of each cell in the grid is calculated by checking if any part of the object or agent is inside the cell. This is be done using a binary function $B\left(x\right)$ that returns 1 if x is inside the object and 0 otherwise. The occupancy $O_i$ of the i-th cell can be calculated as:
\[o_i=max_{j\in O}B\left(c_i-p_j\right)\] 
Where:

 $O$ is the set of all objects in the environment.

 $P_i$ is a point on the surface of the i-th object.

 $B\left(C_i-P_i\right)$ is a binary function that returns 1 if x is inside the object and 0 otherwise.

\subsubsection{Feature Calculation: }
In addition to occupancy, we can also calculate other features of each cell, such as distance to the nearest object or the average color of objects inside the cell. The feature value \$f\_$\mathrm{\{}$i,k$\mathrm{\}}$\$ of the k-th feature in the i-th cell can be calculated as:

\[f_{i,k}=n\sum_{j\in O}{B\left(c_i-p_j\right)f_{j,k}}\] 
Where:

 $O$ is the set of all objects in the environment.

 $P_i$ is a point on the surface of the i-th object.

 $B\left(C_i-P_i\right)$ is a binary function that returns 1 if x is inside the object and 0 otherwise.

 $n$ is the number of objects in the environment.

 \subsubsection{Frontal Shape Observation }

 The Frontal Shape Sensor is a type of visual sensor that captures the shape and appearance of objects in the environment from a frontal perspective. The sensor works by capturing an image of the environment from the agent's current position and angle, and then processing the image to detect and classify objects based on their shape and appearance.

 \begin{center}
 
\begin{tabular}{|p{1.7in}|p{3in}|} \hline 
\textbf{ {Debug View}} & \textbf{ {Grid Buffer}} \\ \hline 
 {\includegraphics*[width=1.84in, height=1.87in]{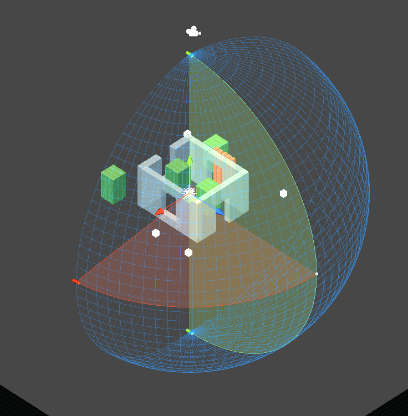}\newline  {Figure 12 3D Debug View} } &  {\includegraphics*[width=2.97in, height=1.84in]{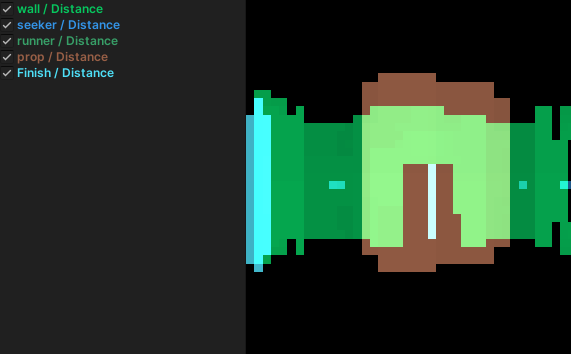}\newline  {Figure 13 2D Grid Buffer}} \\ \hline 
\end{tabular}
\end{center}

\par
The frontal field of vision is responsible for getting the position and shape of detectable objects Infront of the viewing agent. The initial collider buffer is set to 400 but in case agents detect more; the buffer will double itself. The latitude angle north and south are set to 90 degrees each while the longitudinal angle is set to 84 degrees yielding a smaller field of vision. The arc angle of a single FOV grid cell in degrees. Determines the sensor resolution:
\[C_d=\ \pi *2*D\frac{360}{C_d}\] 
Where Cd is equal to cell size at distance and Ca equals cell arc value.

The following is the terminology used to set values in the sensors:

\textbf{Lat Angle North} - The FOV's northern latitude (up) angle in degrees.

\textbf{Lat Angle South} - The FOV's southern latitude (down) angle in degrees.

\textbf{Lon Angle} - The FOV's longitude (left \& right) angle in degrees.

\textbf{Min Distance} - The lowest possible detection distance (near clipping).

\textbf{Max Distance} - The upper limit detection distance (far clipping).

\textbf{Normalization} - How to normalize object distances. 1 for linear normalization. Set the value to $\mathrm{<}$ 1 if observing distance changes at close range is more critical to agents than what happens farther away.

\subsubsection{Spatial Position Observation}
The Spatial Sensor can be used to detect the presence of specific objects in the environment and can provide information about the relative location and orientation of those objects with respect to the agent. The Spatial field of vision is the surround positioning lidar used to identify position as well as the distance to the surrounding detectable objects its initial collider buffer is set to 32 and if agents detect more the buffer will double itself. Latitude angles north and south are set to 90 degrees each while the longitudinal angle is set to 180 degrees yielding a 360 field of vision. and 2D ray casts pointing outward along the x and z-axis around the agent and agents' velocity with its facing direction along the z-axis.

\begin{center}
\begin{tabular}{|p{2.1in}|p{2.3in}|} \hline 
\textbf{ {Debug View}} & \textbf{ {Grid Buffer}} {} \\ \hline 
 {\includegraphics*[width=1.97in, height=1.90in]{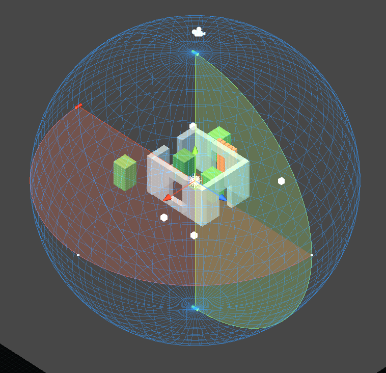}\newline  {Figure 14 3D Debug View} } &  {\includegraphics*[width=2.19in, height=1.87in, trim=2.70in 0.00in 1.59in 0.00in]{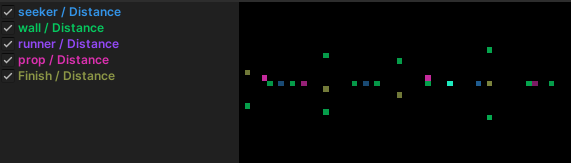}\newline  {Figure 15 2D Drif Buffer}  } \\ \hline 
\end{tabular}
\end{center}
 
The Spatial Sensor represents the environment as a set of spatial features. This sensor captures information about the location and characteristics of objects in the environment, such as their position, orientation, and size.

\subsubsection{Ray-casts Observation}
Agents have 2D Ray-casts that surround them observing. In ML-Agents, Ray-cast Sensors are a type of sensor that provides information about the environment to the agent. A Ray-cast Sensor works by casting a ray or multiple rays from a point on the agent's body to detect objects in the environment. The sensor returns information about the distance, angle, and type of the detected objects. Ray- cast Sensors are commonly used in robotics and game development to simulate perception and enable agents to interact with the environment. In ML agents, Ray-cast Sensors can be used to provide the agent with information about the environment, such as the location of obstacles, the distance to objects, and the presence of other agents.

\begin{center}
\includegraphics*[width=3.65in, height=3.27in]{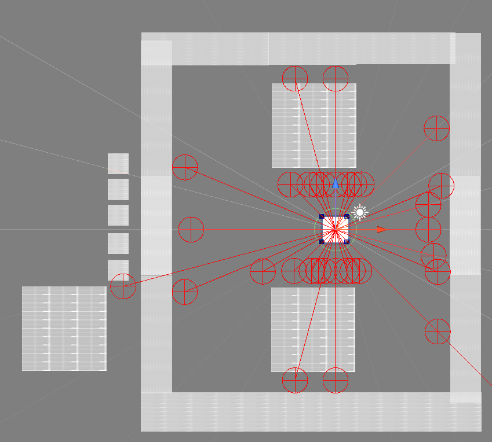}
  {Figure 16 2d Ray casts}
\end{center} 
The agent has 8 ray casts per direction with a max ray degree of 180 which means by adding both directions we get a 360 view. The sphere ray cast radius is 0.3 while the ray length is set to 20.

\subsubsection{Realtime values via script}
Values include normalized potion of self, normalized velocity, facing direction vector, normalized rotation, a bool telling if the agent is dragging prop, and a normalized timer that finishes when environmental steps finish.
\newpage

\subsection{State Diagram:}
\begin{center} 
 \textbf{}
 \includegraphics*[width=6.02in, height=5.53in]{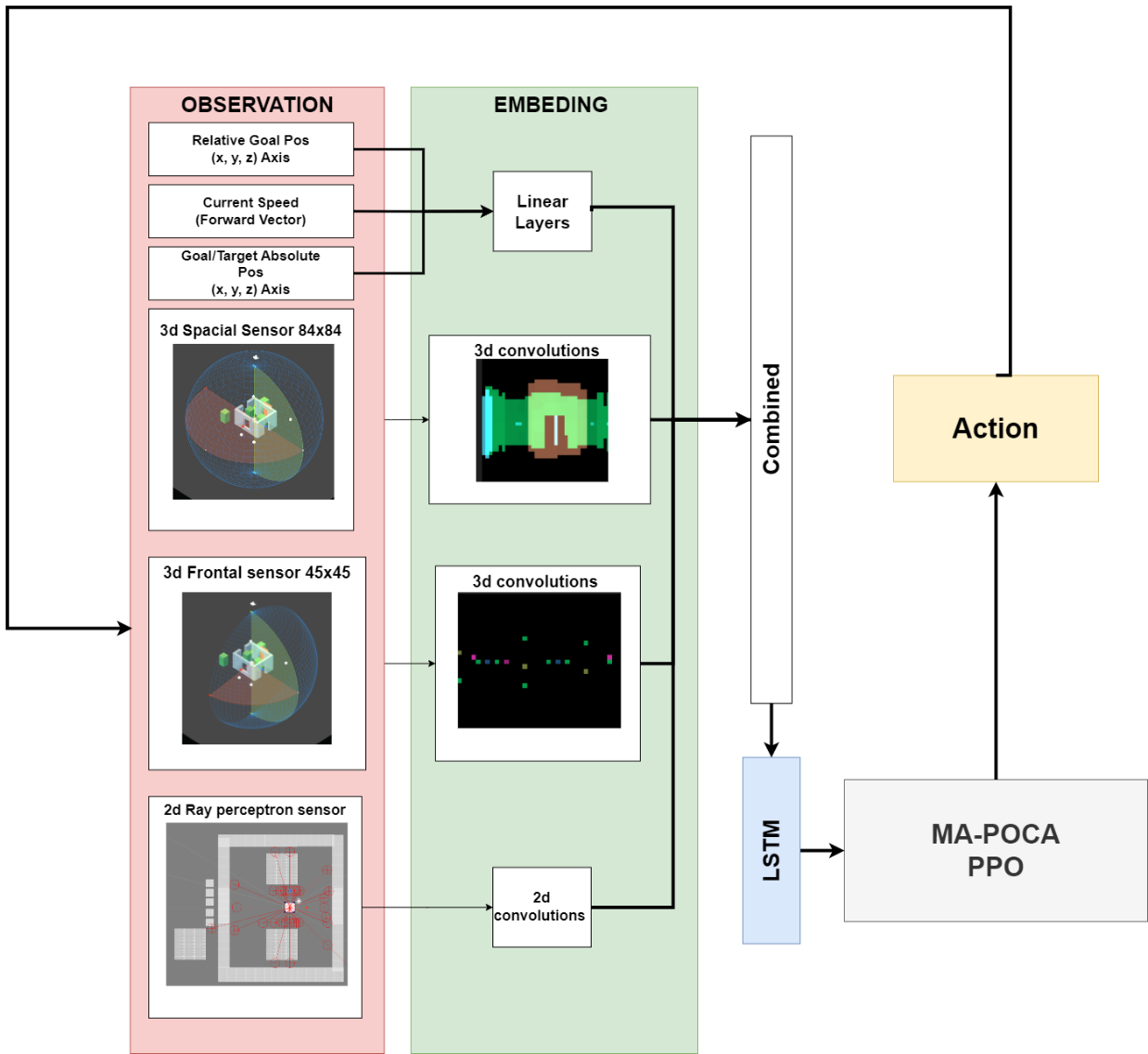}
  {Figure 17 State Diagram of Agents observation to action flow}
\end{center}

Figure 1 represents the observation being provided to the agents to generate actions and develop the best policies. The observations are then fed into MA-Poca for Seekers and PPO for Hiders.

Observations including relative position, current velocity, target relative position, 3d spatial sensor, 3d frontal sensor which includes 84x84 RGB camera, and 2d ray-casts are taken from the agent to the observation buffer and sent to generate optimal policy to generate the best result. 
\newpage

\subsection{ Pre-Hider and Seeker Experiments}
 
We conducted three different yet progressive experiments including the "hummingbird experiment", "target drone experiment" and the "eye experiment".

\begin{center} 
\begin{tabular}{|p{1.6in}|p{1.2in}|p{1.2in}|} \hline 
 {Hummingbird} &  {Target Drone} &  {The Eye} \\ \hline 
 {\includegraphics*[width=1.39in, height=1.21in]{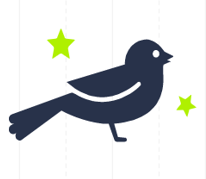}} &  {\includegraphics*[width=1.29in, height=1.12in]{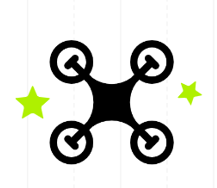}} &  {\includegraphics*[width=1.40in, height=1.14in]{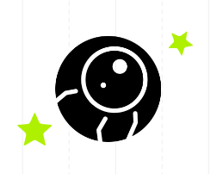}} \\ \hline 
\textbf{ {Info}} {:\newline Training a humming bot to collect nectar [14]} & \textbf{ {Info}} {:\newline Training a drone bot to tag target} & \textbf{ {Info}} {:\newline Training an eye bot to avoid collision with ally and reach target} \\ \hline 
\textbf{ {Conditions:\newline }} {Stationary Target\newline Wide boundaries\newline Known Target\newline } & \textbf{ {Conditions:\newline }} {Dynamic Target\newline Normal boundaries\newline Known Target\newline } & \textbf{ {Conditions:\newline }} {Dynamic Target\newline Small boundaries\newline unknown Target\newline } \\ \hline 
\end{tabular}
\end{center} 
\textbf{}

\subsubsection{ Hummingbird Agent - Experiment}
Observing agent behavior, the Agent seems to wobble around trying different actions to be able to achieve reward and avoid punishment.

\textbf{Reward} = +.01f if Agent is in Nectar

\textbf{Reward} = -0.5 if Agent Collides

\includegraphics*[width=6.00in, height=2.42in]{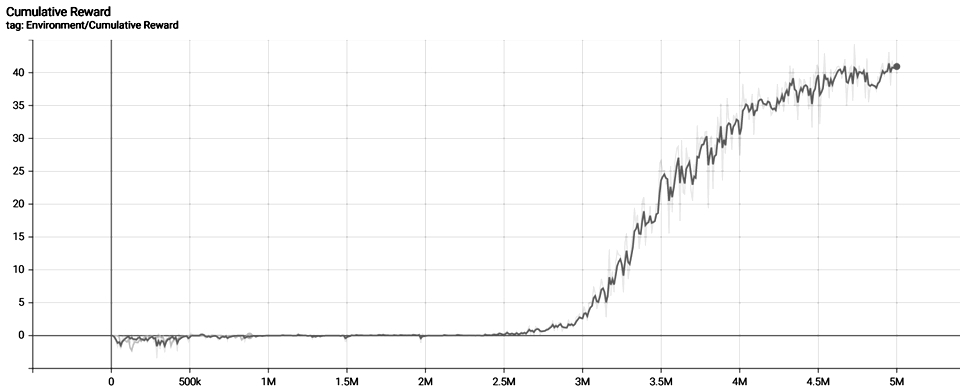}
\begin{center}
  {Figure 18 Cumulative Reward for ``Hummingbird'' agent}
\end{center}  
 
After approx. 2.5M Steps, Agent seems to figure out a pattern to successfully navigate to the nectar Collider.

\subsubsection{Drone Agent Experiment }
Observing Agents Behavior, we find Drone Agent seems to wobble around trying different actions to be able to achieve reward and avoid punishment.

\textbf{Reward} = +1f if Agent is in Nectar

\textbf{Reward} = -1 if Agent Collides

\includegraphics*[width=6.00in, height=2.04in]{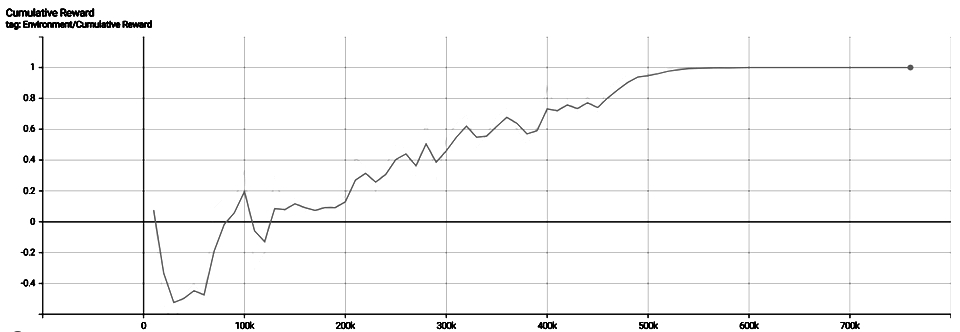}
\begin{center}
  {Figure 19 Cumulative Reward for "Drone Target" agent}
\end{center} 
 
\subsubsection{EyeAgent Experiment}

Observing agent behavior, the Agent seems to wobble around trying different actions to be able to achieve reward and avoid punishment.

\textbf{Reward} = +1f if Agent is in Target

\textbf{Reward} = +0.001 if Agent Looks at Target

\textbf{Reward} = -0.2 if Agent Collides

\textbf{Reward} = -0.5 if Agent Collides

\includegraphics*[width=5.98in, height=3.89in]{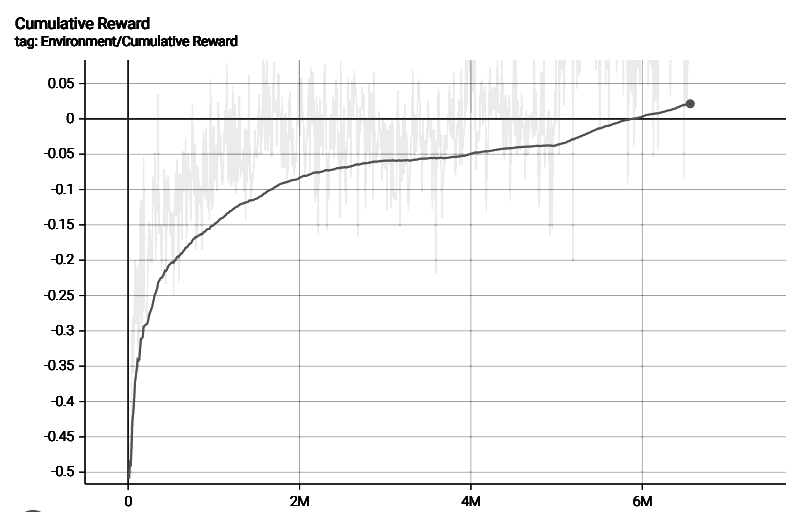}
\begin{center} 
  {Figure 20 Cumulative Reward for "The Eye" agent}
\end{center} 

After approx. 6.5M Steps, The Agent seems to figure out a pattern to successfully stop colliding with walls and with other agents.

\subsection{Instance Setup}
For our research, we are conducting two separate training sequentially. Seekers are using MA-POCA, which is used to train a group of seekers (1-4). While the hider agents are using PPO (Proximal Policy Optimization) which is used to train hider agents. They use the props available in the environment to shield themselves for a shorter reward or use it to block the windows and doorways (best policy).

\begin{center}
\includegraphics*[width=4.34in, height=3.22in]{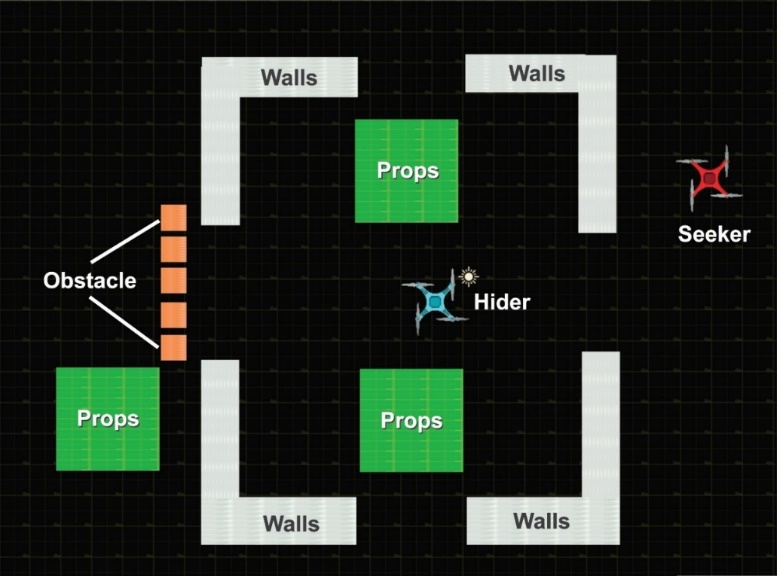}
{Figure 21 Environment visual (2d view)}
\end{center}

Figure 21 demonstrates the training and testing environment that consists of 1-2 Hiders Drones, 1-4 Seeker Drones, 1-4 Props (doors, windows), 5 obstacles, 4 ``L-shaped'' walls, and 4 boundaries around the whole setup.

\textbf{Seeker}: (MA-POCA) The Hunter Drone Agent is tasked to locate and hit Hider to avoid walls and obstacles.

\textbf{Hider}: (PPO) The Escaping Drone Agent tasked to Lock doors and clear obstacles.

\textbf{Props}: Draggable/Lockable entity used by agents for their gain.

\textbf{Walls}: Static non-movable entity used to teach navigation.

\textbf{Obstacle}: Physics-based Rigid bodies blocking agents' path to the desired goal.
\newpage

\subsubsection{Parallelism}
Training is performed using the principle of parallelism. ``n'' consecutive similar environment prefabs are initialized altogether. Each has an identical list of agents and has a similar environment structure. Experience can refer to the dataset of observations and rewards used to train an agent's neural network. Let $E$ be the experience in n instances:
\[E_t=\ \sum^n_{n=1}{(E_1+\ E_2}+\ E_3\ .\ .\ .\ .\ .\ .\ E_n\ )\ \] 

\begin{center}
\includegraphics*[width=4.17in, height=3.47in]{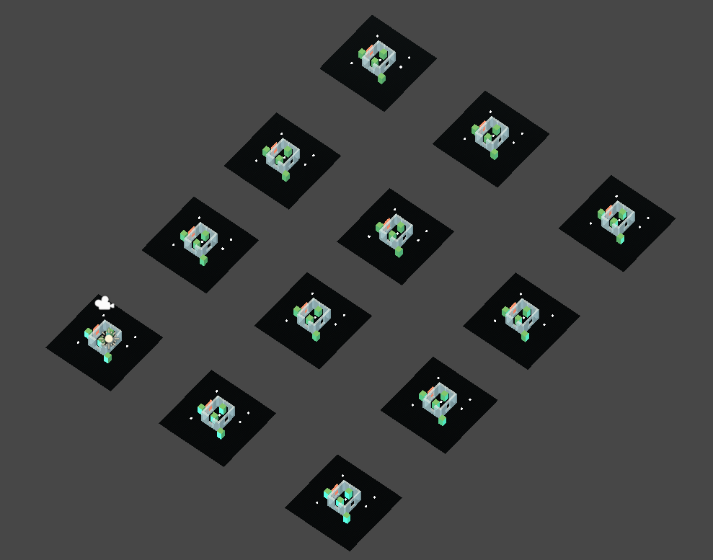}
\\
{Figure 22 Parallel Environment visual (3D view)}
\end{center} 
 
By training multiple agents at once, the overall training time can be reduced, as the agents can learn from their experiences in parallel rather than one at a time. It is achieved using a technique called asynchronous training, where multiple agents are trained concurrently, and their experiences are used to update the neural network at different times. 

Figure 13 demonstrates Trajectory-based parallelism which involves running multiple instances of the simulation environment in parallel, with each instance running a different agent. Each agent collects a sequence of experiences, or "trajectory", from interacting with the environment, and these trajectories are used to update the agent's model.

\subsubsection{Multi-Agent Prep-Phase}
Agents are given a total of 3072 steps in which they can make decisions every third step and update their actions accordingly. Hiders are given a 40\% preparation time before the seekers are allowed to move. In this ``prep phase'' hiders are allowed to move freely while avoiding contact with seekers. Hiders can learn to identify, drag, and lock movable props (i.e., boxes) to use them to their advantage. During this period, seekers are observing the hider's location and shapes (if hiders accidentally stay Infront of the seeker's field of vision) but they are restricted with their movement.
\[S_p=\ S_m*\left(\frac{40}{100}\right)\] 
Where $S_p$ are the Steps for the prep phase and $S_m$ are the Max Steps.

\begin{tabular}{|p{1.1in}|p{0.6in}|p{1.1in}|p{0.6in}|p{0.6in}|} \hline 
 {prep\_phase\_steps} &  {=} &  {Max\_env\_steps} &  {*} &  {(40/100)} \\ \hline 
 {1228.8} &  {$\sim$} &  {3072} &  {*} &  {(40/100)} \\ \hline 
\end{tabular}

\newpage

\subsubsection{Multi-Agent Test-Phase}
The test phase refers to that portion of training in which the seekers are let loose and given control over themselves. That means seekers can now also move along with hider agents and try to hunt them and tag them. After the 40\% steps are completed out of max environment steps, for the remaining 60\% steps, seekers are allowed to move and use their learned policy to focus on targets and reach them.
\[S_t=\ S_m*\left(\frac{60}{100}\right)\] 
Where $S_t$ are the Steps for the test phase and $S_m$ are the Max Steps.

\begin{tabular}{|p{1.1in}|p{0.6in}|p{1.1in}|p{0.6in}|p{0.6in}|} \hline 
 {test\_phase\_steps} &  {=} &  {max\_env\_steps} &  {*} &  {(60 / 100)} \\ \hline 
 {1843.2} &  {$\sim$} &  {3072} &  {*} &  {(60 / 100)} \\ \hline 
\end{tabular}

\subsection{PLAYTESTING RL AGENTS}
\subsubsection{ Observing Player Progression via Curriculum learning}
Our Hider agents are trained via a form of machine learning method known as curriculum learning which includes progressively raising the level of complexity of the training examples provided to an AI agent. Task Sequencing: The first step in curriculum learning is to define a sequence of tasks with increasing difficulty. This sequence can be represented as a function that maps the current training iteration to the corresponding task. For example, the task at iteration $t$ could be represented as:
\[f\left(t\right)=Task_t\] 
The plan is to start with simple instances and progressively get more complicated as the agent gets better at the task.
\[C_r>\ T_r\to D\ +=1\] 
Where $C_r$ is the current reward, $C_r$is the Cumulative threshold reward. $D$ is the difficulty as in levels. If the current reward exceeds a given number of episodes, the environment will evolve to complex itself to make it harder for an agent to learn gradually. Agents are introduced with four different levels of environmental setup in Figure 12. Levels are made difficult to complete progressively. For example, level one has the least difficulty while level four has the maximum.  Curriculum learning is applied to educate the agents on how to play the game of hide and seek more effectively. The hiders and the seekers are two different teams of agents in this game. The hiders must conceal themselves in their surroundings, while the seekers must track them down and tag them. By gradually escalating the level of difficulty, curriculum learning is integrated into the game of hide and seek with the students. The hiding places could be harder to locate as the hiders get more adept at doing so. Similar to this, as the game goes The curriculum learning in ml-agents hide and seek is intended to assist the agents in learning more quickly and effectively. The agents can learn the game in a more organized and effective manner by starting with simple instances and progressively adding complexity.
 
\newpage
  
\subsubsection{Seeker's Assault Horizon}
The idea behind this is inspired by the military term "Dog Fight". A dogfight is a kind of aerial conflict in which two or more aircraft engage in a close-quarters battle. High-speed maneuvers and intricate strategies are frequently used, with each pilot attempting to outwit their rival. Dogfights may be quite dangerous since they frequently take place at great heights and make use of cutting-edge equipment. Our agent's navigational capabilities are designed by keeping the "running and chasing" ability in mind. Seekers are to keep hiders in their assault horizon to gain rewards and even tag them to get additional rewards. 

 With the help of a specially designed target field of vision ``reward strength'' signal, seekers can identify what angle suits best to adjust its rotation and position accordingly, so it is facing and is right Infront of its target.

Let $T$ = forward vector ``z-axis'' i.e. (0,0,1) and let $a$ = difference between this agent's position  $P_1$ and other agent's position $P_2$.
\[a=\ P_2-P_1\] 
\[R_{d\ }=\ \left|a\right|\] 
\[d=T.\left(\frac{a}{R_d}\right)\] 
\[S_s=\left(\frac{d}{R_d}\right)\] 
Where $S_s$ is the Signal Strength, $R_{d\ }$is the magnitude of $a$, { }$d$ represents dot product between forward vector and $\left(\frac{a}{R_d}\right)$

\begin{center} 
\begin{tabular}{|p{2.2in}|p{1.9in}|} \hline 
\textbf{ {Top View}} & \textbf{ {Side View}} \\ \hline 
{\includegraphics*[width=2.14in, height=1.82in]{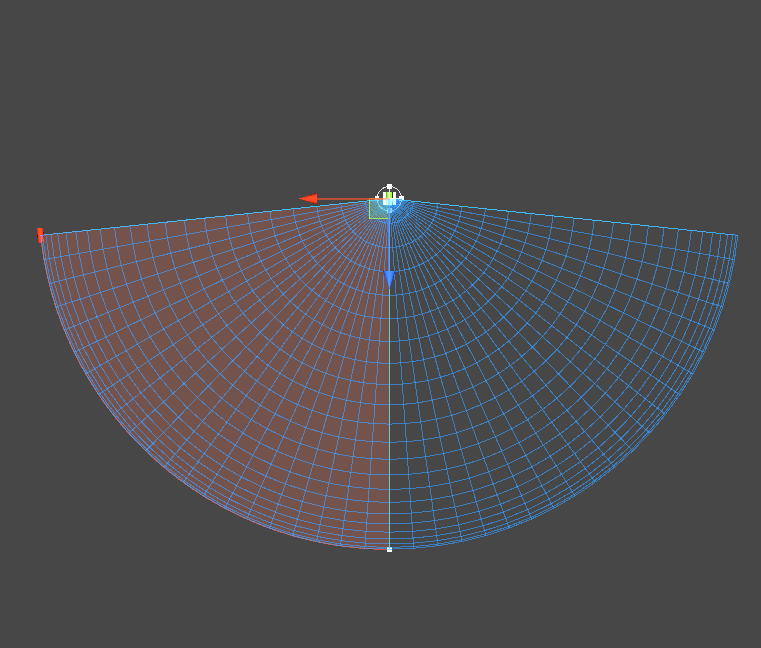}\newline  {Figure 23 Top View} } &  {\includegraphics*[width=1.88in, height=1.82in]{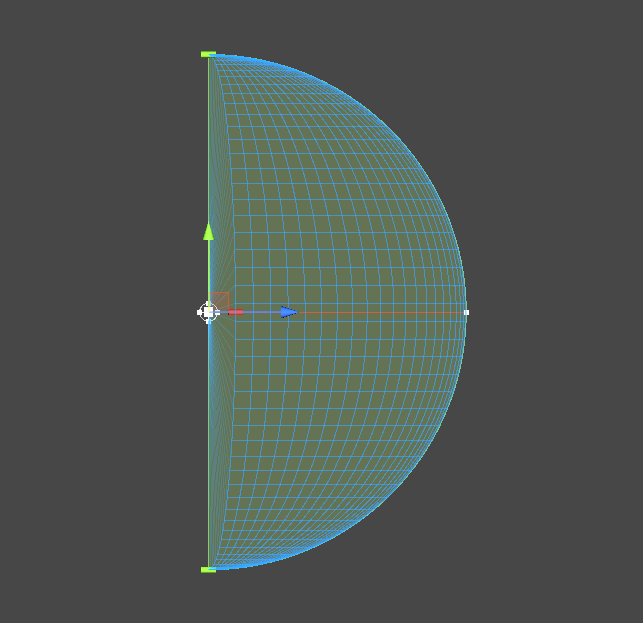}\newline {Figure 24 Side View}} \\ \hline 
\end{tabular}
\end{center} 
\textbf{}

\newpage 

\subsubsection{ Hider's Prop Use}
As our agents have flying-like capabilities, introducing a ramp was of no use as agents might learn to just fly over the wall. So we introduced windows and doorways which when blocked with props, make a safe house for the hiders. There is a single prop assigned for each window or door. Agent must learn to efficiently learn to handle the props to minimize the time required to close that door and also learn which prop is best for quickly closing a specific location door. If the agent is using the farthest prop to close a door It will damage the efficiency as it needs to close the rest of the doors too.

\begin{center}
\begin{tabular}{|p{2.1in}|p{2.2in}|} \hline 

{\includegraphics*[width=2in, height=2in]{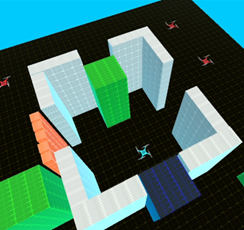}\newline }Figure 25 Prop 1 used successfully\textbf{} & 

{\includegraphics*[width=2in, height=2in]{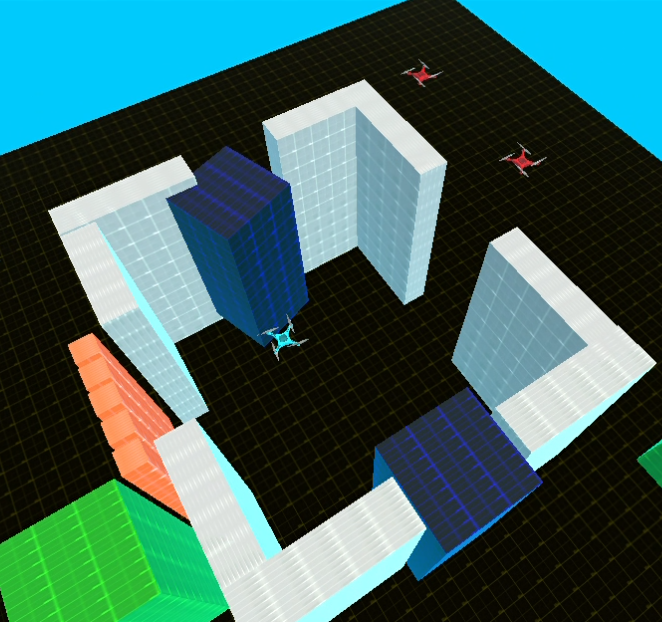}\newline }Figure 26 Prop 2 used successfully.\newline \textbf{ {}} \\ \hline 

{\includegraphics*[width=2in, height=2in]{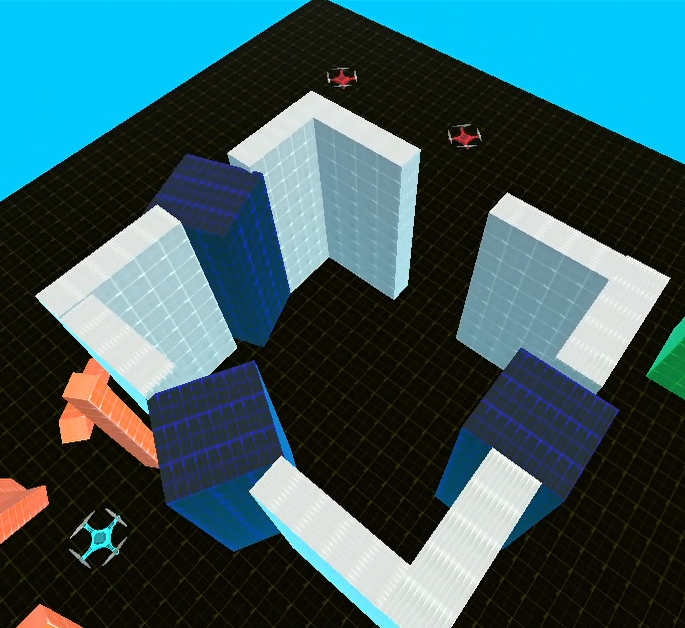}\newline }Figure 27 Prop 3 used successfully while Avoiding obstacles.\textbf{} &  {\includegraphics*[width=2in, height=2in, trim=0.00in 0.09in 0.00in 0.00in]{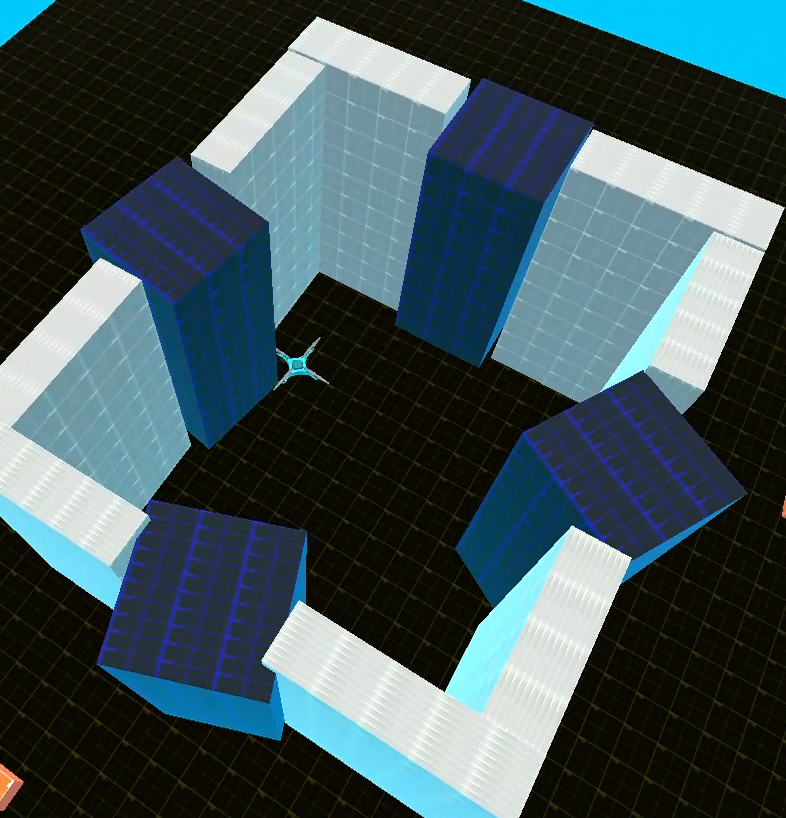}\newline }Figure 28 Prop 4 with max distance and avoiding seekers collision used successfully.\newline \textbf{ {}} \\ \hline 
\end{tabular}
\end{center}
 
\newpage

\subsubsection{ Clearing Obstacles}
\begin{center}
\begin{center}
    \includegraphics*[width=2.55in, height=2.78in]{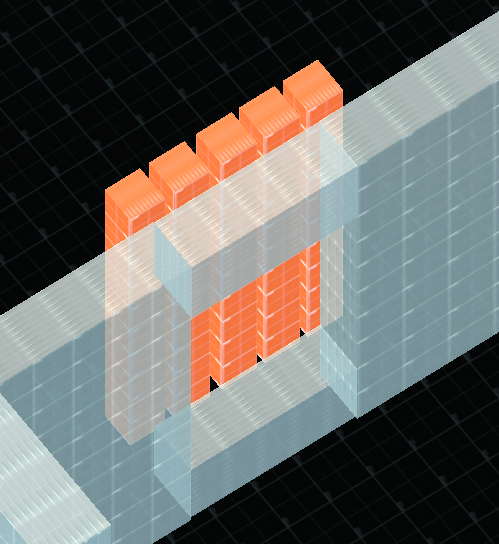} 
    
\end{center}

  {Figure 29 Obstacles are demonstrated with orange color.}
\end{center}
 
Agents have an additional problem of clearing the obstacles to make a path for closing the doorway. These orange blocks have mass and light gravity. This light gravity increases the difficulty as it's more complex to clear them away with one stroke as they start to float in the path of agents. Agents need to constantly move away from the obstacles and clear their path to optimal solutions.

\end{document}